\documentclass[10pt, a4paper]{article}
\usepackage{lrec}
\usepackage{graphicx}
\usepackage{tabularx}
\usepackage{soul}

\usepackage{epstopdf}
\usepackage[utf8]{inputenc}

\usepackage{hyperref}
\usepackage{xstring}
\usepackage{booktabs}
\usepackage{linguex}
\usepackage{longtable}
\usepackage{float}

\title{DiscSense: Automated Semantic Analysis of Discourse Markers}

\name{Damien Sileo$^{3,1}$, Tim Van de Cruys$^{2}$, Camille Pradel$^{1}$, Philippe Muller$^{3}$}

\address{1:Synapse D\'eveloppement, 2:IRIT (CNRS), 3:IRIT (University of Toulouse) \\
         damien.sileo@irit.fr\\}

\abstract{
Discourse markers ({\it by contrast}, {\it happily}, etc.) are words
or phrases that are used to signal semantic and/or pragmatic
relationships between clauses or sentences. Recent work has fruitfully
explored the prediction of discourse markers between sentence pairs in
order to learn accurate sentence representations, that are useful in
various classification tasks. In this work, we take another
 perspective: using a model trained to predict
discourse markers between sentence pairs, we predict plausible markers
between sentence pairs with a known semantic relation (provided by
existing classification datasets). These predictions allow us to study
the link between discourse markers and the semantic relations annotated in classification datasets. 
Handcrafted mappings have been proposed between markers and discourse relations on a limited set of markers and a limited set of categories, but there exist hundreds of discourse markers expressing a wide variety of
relations, and there is no consensus on the taxonomy of relations
between competing discourse theories (which are largely built in a
top-down fashion). 
By using an automatic prediction method over
existing semantically annotated datasets, we provide a bottom-up
characterization of discourse markers in English.
The resulting dataset, named DiscSense, is publicly available.
\\ \newline  }

\begin{document}

\maketitleabstract

\section{Motivation}

Discourse markers are a common language device used to make explicit the semantic and/or pragmatic relationships between clauses or sentences. For example, the marker \textit{so} in
sentence~\ref{ex:gasoline} indicates that the second clause is a consequence of the first.

\ex. \label{ex:gasoline} We're standing in gasoline, so you should not
smoke.

Several resources enumerate discourse markers and their use in
different languages, either in discourse marker lexicons
\cite{DBLP:phd/ethos/Knott96,Stede:02,roze12discours,das-etal-2018-constructing} or in corpora, annotated with discourse
relations, such as the well-known English Penn Discourse TreeBank
\cite{pdtb2.0}, which inspired other efforts in 
Turkish, Chinese and French
\cite{zeyrek:discourse:2008,zhou:chinese:2014,danlos:hal-01392807}.
The PDTB identifies different types of discourse relation categories (such as
{\it conjunction} and {\it contrast}) and the respective markers that
frequently instantiate these categories (such as {\it and} and {\it
  however}, respectively), and organizes them in a three-level
hierarchy.
It must be noted, however, that there is no general consensus on the
typology of these markers and their rhetorical functions. As such,
theoretical alternatives to the PDTB exist, such as Rhetorical Structure Theory or RST
\cite{DBLP:conf/sigdial/CarlsonMO01}, and Segmented Discourse Representation Theory or SDRT
\cite{asher2003logics}. Moreover, marker inventories focus on a restricted number of rhetorical relations that are too coarse and not exhaustive, since discourse marker use depends on the grammatical, stylistic, pragmatic, semantic and emotional contexts that can undergo fine grained categorizations.

Meanwhile, there exist a number of NLP classification tasks (with
associated datasets) that equally consider the relationship between
sentences or clauses, but with relations that possibly
go beyond the usual discourse relations; these tasks focus on various phenomena such as
implication and contradiction \cite{bowman2015large},
semantic similarity, or paraphrase
\cite{DBLP:conf/coling/DolanQB04}.  Furthermore, a number of tasks
consider single sentence phenomena, such as sentiment, subjectivity,
and style. Such characteristics have been somewhat ignored for the
linguistic analysis and categorization of discourse markers {\it per se}, even though discourse markers have been successfully used to improve categorization performance for these tasks
\cite{Jernite2017,Nie2017,pan-etal-2018-discourse,sileo2019discovery}.
Specifically, the afore-mentioned research shows that the prediction
of discourse markers between pairs of sentences can be exploited as a
training signal that improves performance on existing classification
datasets.  In this work, we make use of a model trained on discourse
marker prediction in order to predict plausible discourse markers
between sentence pairs from existing datasets, which are annotated
with the correct semantic categories. This allows us to explore the
following questions:

\begin{itemize}
 \item[--] Which semantic categories are applicable to a particular
   discourse marker (e.g. is a marker like {\it but} associated with
   other semantic categories than just mere contrast)?
 \item[--] Which discourse markers can be associated with the semantic
   categories of different datasets (e.g. what are the most likely
   markers between two paraphrases)?
 \item[--] To what extent do discourse markers differ between datasets
   with comparable semantic categories (e.g. for two sentiment
   analysis datasets, one on films and one on product reviews, are the
   markers associated with positive sentences different)?
\end{itemize}

\begin{figure*}[]
\centering

\includegraphics[width =0.78\textwidth]{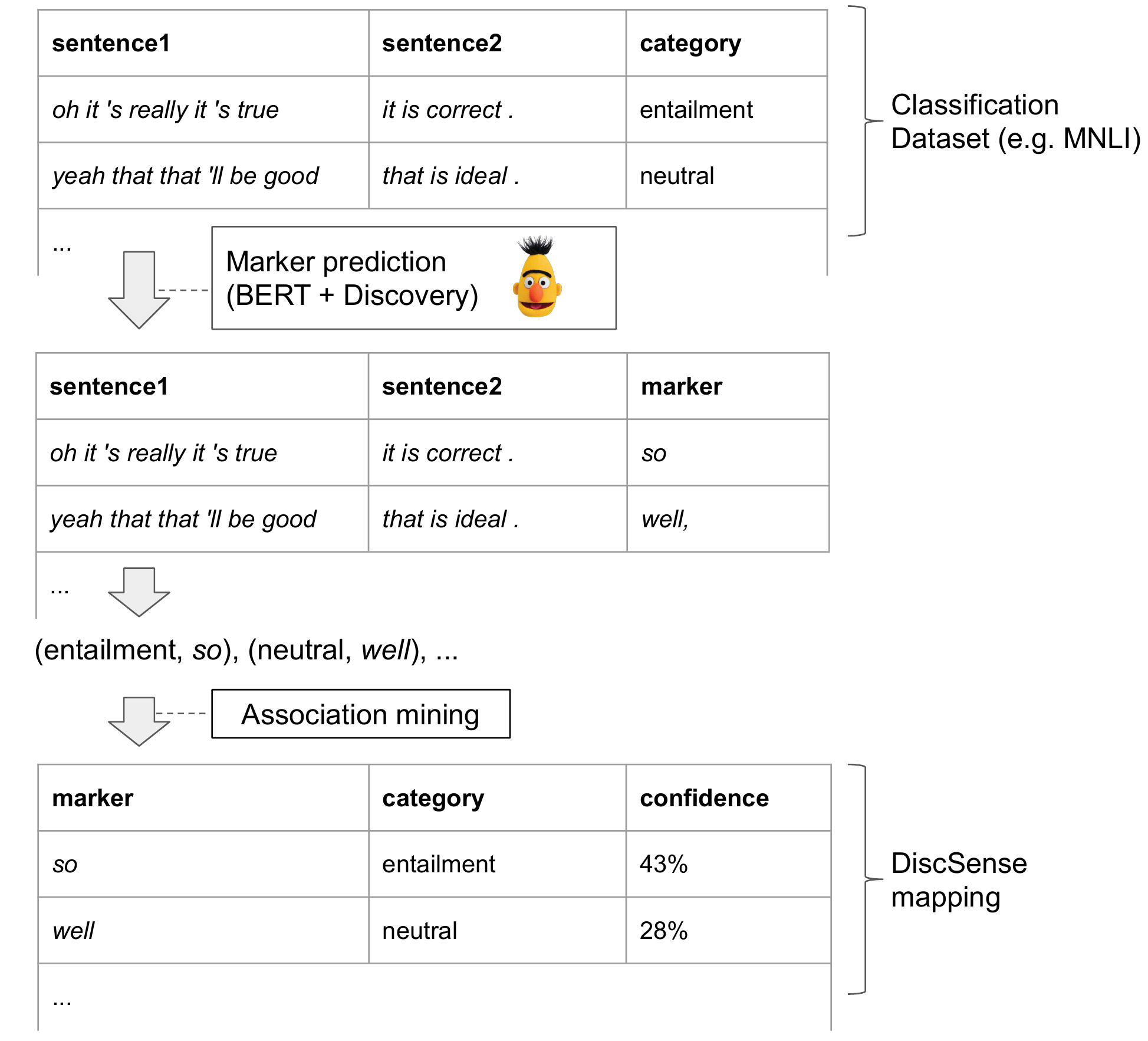}
  \caption{Overview of our method}
\label{fig:overview} 
\end{figure*}

In order to answer these questions, we train a model for
discourse marker prediction between sentence pairs, using millions of
examples.  We then use this model to predict markers between sentences
whose semantic relationships have already been annotated---for example,
pairs of sentences ($s_1$,$s_2$,$y$) where $y$ is in {\it Paraphrase},
{\it Non-Paraphrase}. These predictions allow us to examine the
relationship between each category $y$ and the discourse markers that
are most often predicted for that category. Figure \ref{fig:overview} shows an overview of our method.

%

 Thus, we propose {\it DiscSense}, a mapping between markers and senses, that has several applications:
 \begin{itemize}
  \item[--] A characterization of discourse markers with categories that provides new knowledge about the connotation of discourse markers; our characterization is arguably richer since it does not only use PDTB categories. For instance, our mapping shows that the use of some markers is associated with negative sentiment or sarcasm; this might be useful in writing-aid contexts, or as a resource for second language learners; it could also be used to guide linguistic analyses of markers; 
  \item[--] A  characterization of categories of discourse markers can help ``diagnosing" a classification dataset; 
  As shown in table \ref{tab:associations} below, SICK/MNLI dataset categories have different associations and our method can provide a sanity check for annotations (e.g. a Contradiction class should be mapped to markers expected to denote a contradiction);
 \item[--]  An explanation of why it is useful to employ discourse marker prediction as a training signal for sentence representation learning; DiscSense can also be used to find markers which could be most useful when using a discourse marker prediction task as auxiliary data in order to solve a given target task. 
\end{itemize}


\section{Related work}

Previous work has amply explored the link between discourse markers
and semantic categories.
\newcite{pitler2008}, for example, use the PDTB to
analyze to what extent discourse markers {\it a priori} reflect
relationship category. \newcite{Asr2012ImplicitnessOD} have
demonstrated that particular relationship categories give rise to more
or less presence of discourse markers. And a recent categorization of
discourse markers for English is provided in the DimLex lexicon
\cite{das-etal-2018-constructing}.

As mentioned before, discourse markers have equally been used as a learning signal for the
prediction of implicit discourse relations
\cite{Liu2016ImplicitDR,braud-denis:2016:EMNLP2016} and inference
relations \cite{Pan2018DiscourseMA}. This work has been generalized by
DiscSent \cite{Jernite2017}, DisSent \cite{Nie2017}, and Discovery
\cite{sileo2019discovery} who use discourse markers to learn general
representations of sentences, which are transferable to various NLP
classification tasks. However, none of these examine the individual
impact of markers on these tasks.


\section{Experimental setup}

\subsection{Discourse marker corpus}


In order to train a model to predict plausible discourse markers
between sentence pairs, we use the English {\it Discovery}
corpus \cite{sileo2019discovery}, as it has the richest set of markers. It is
composed of 174 discourse markers with 20{\sc k} usage examples for each marker
(sentence pairs where the second sentence begins by a given marker).  Sentence pairs were extracted from web data
\cite{Panchenko2017}, and the markers come either from the PDTB or
from an automatic extraction method based on heuristics. An example of
the dataset is provided in \ref{ex:discovery}.

\exg. {Which is best?} {Undoubtedly,} {that depends on the person.}\\
$s_1$ $c$           $s_2$  \\ \label{ex:discovery}

Since we plan to use marker prediction on sentence pairs from classification
datasets, in which some sentence pairs cannot plausibly occur
consecutively, (e.g. entirely unrelated sentences),
we augment the {\it Discovery} dataset with non-consecutive sentence
pairs from the DepCC corpus for which we create a new class. We sample sentences that were separated by 2 to 100
sentences in order to cover various degrees of relatedness.  

Besides, we also want to predict markers beginning single
sentences, so we mask the first sentence of {\it Discovery} example pairs  in 10\%
of cases by replacing it with a placeholder symbol $[S_1]$. This placeholder will  be used to generate sentence pairs from single sentence in datasets where sentence pairs are not available. For example, in the Customer Review dataset (CR), we predict a marker between $[S_1]$ and review sentences.

In addition, we also use another dataset by \newcite{Malmi2018} for which human annotator accuracy is available for a better assessment of the performance of our marker prediction model. It
contains 20{\sc k} usage examples for 20 markers extracted from
Wikipedia articles (the 20 markers are a subset of the markers
considered in the {\it Discovery} dataset); we call this dataset {\it
  Wiki20}.

\begin{table}[htb]
\footnotesize
\hskip-0.3cm
\centering
\begin{tabular}{lll}
\toprule
                        & Wiki20 & Discovery \\
 \midrule
Majority Class                & 5.0      & 0.6      \\
Human Raters            & 23.1  & -         \\
Decomposable Attention     & 31.8   & -         \\
Bi-LSTM & -      & 22.2      \\
\midrule
BERT+Discovery          & 30.6     & \textbf{32.9}      \\
BERT+Discovery+Wiki20   &\textbf{47.6}     & -   \\     
\bottomrule
\end{tabular}
\caption[Discourse marker prediction accuracy]{Discourse marker prediction accuracy percentages on {\it Wiki20} and {\it Discovery} datasets. Human Raters and Decomposable Attention are from \cite{Malmi2018}. Bi-LSTM is from \cite{sileo2019discovery} and the last two are ours.}
\label{tab:conn-pred}
\end{table}
\subsection{Classification datasets}
We leverage classification datasets from DiscEval \cite{sileo2019discoursebased}, alongside GLUE classification tasks \cite{wang2018glue} augmented with SUBJ, CR and SICK tasks from SentEval \cite{conneau-kiela-2018-senteval} in order to have different domains for sentiment analysis and NLI. We map the semantic similarity estimation task (STS) from GLUE/SentEval  into a classification task by casting the ratings into three quantiles and discarding the middle quantile.
Table \ref{tab:classification-datasets} enumerates the classification datasets we used in our study.

\begin{table}[htb]
\begin{footnotesize}
\centering

\begin{tabular}{p{0.73in}l@{\hskip -0.0in}r@{\hskip +0.05in}r}
\toprule
marker & category &  support & confidence (prior)
     \\
\midrule
   unfortunately, &                                      CR.negative &       66 &     100.0 (21.8) \\
           sadly, &                                      CR.negative &       20 &      95.2 (21.8) \\
   unfortunately, &                                   SST-2.negative &      240 &      96.0 (22.5) \\
     as a result, &                                   SST-2.negative &       65 &      94.2 (22.5) \\
     in contrast, &                          MNLI.contradiction &     1182 &      74.1 (16.9) \\
       curiously, &                          MNLI.contradiction &     2912 &      70.8 (16.9) \\
     technically, &                         SICKE.contradiction &       29 &       87.9 (7.8) \\
          rather, &                         SICKE.contradiction &      147 &       69.7 (7.8) \\
       similarly, &                             MRPC.paraphrase &       85 &      87.6 (35.5) \\
        likewise, &                             MRPC.paraphrase &      103 &      84.4 (35.5) \\
         instead, &                           PDTB.Alternative &       27 &       22.5 (0.6) \\
            then, &                          PDTB.Asynchronous &       60 &       38.7 (2.4) \\
      previously, &                          PDTB.Asynchronous &       36 &       36.4 (2.4) \\
   by doing this, &                                 PDTB.Cause &       22 &      61.1 (14.8) \\
     additionally &                           PDTB.Conjunction &       47 &      63.5 (12.5) \\
              but &                              PDTB.Contrast &       89 &       61.4 (7.0) \\
       elsewhere, &                                  PDTB.List &       41 &       16.2 (1.3) \\
    specifically, &                           PDTB.Restatement &      100 &      67.6 (10.6) \\
       seriously, &                                SarcasmV2.sarcasm &      225 &      71.2 (26.7) \\
              so, &                                SarcasmV2.sarcasm &       81 &      65.6 (26.7) \\
\bottomrule
\end{tabular}
\caption[Categories and most associated markers]{ 
Sample of categories and most associated markers. CR.neg denotes the negative class in the CR dataset. Datasets are described in table \ref{tab:classification-datasets}.
Support is the number of examples where the marker was predicted given a dataset. Confidence is the estimated probability of the class given the prediction of the marker i.e. $P(y|m)$. The prior is $P(y)$.
A larger version is available in annex A and a full version is available at \url{https://github.com/synapse-developpement/DiscSense}.
}
\label{tab:associations}
\end{footnotesize}

\end{table}

 \begin{table*}
 \footnotesize
 \begin{tabular}{llll}
\toprule
       dataset &                categories &                                                                                 exemple$\&$class & $N_{train}$ \\
\midrule
            MR &         sentiment (movie) &                                                              ``bland but harmless" \hfill neg &       11k \\
           SST &         sentiment (movie) &                                               ``a quiet , pure , elliptical film " \hfill pos &       70k \\
            CR &      sentiment (products) &                                               ``the customer support is pathetic." \hfill neg &        3k \\
          SUBJ &      subjective/objective &                                           ``it is late at night in a foreign land" \hfill obj &       10k \\
          MRPC &               paraphrase  &                                  ``i ’m never going to [...]"/``i am [...]" \hfill paraphrase &        4k \\
        SICK-E &        inference relation &                                          ``a man is puking"/``a man is eating" \hfill neutral &        4k \\
          SNLI &        inference relation &                                             ``dog leaps out"/``a dog jumps" \hfill entailment &      570k \\
     SarcasmV2 &       presence of sarcasm &       ``don't quit your day job"/``[...] i was going to sell this joke. [...]" \hfill sarcasm &        9k \\
      Emergent &                   stance  &                ``a meteorite landed in nicaragua."/``small meteorite hits managua" \hfill for &        2k \\
          PDTB &        discourse relation &                                  ``it was censorship"/``it was outrageous" \hfill Conjunction &       13k \\
       Squinky &                     I/I/F &                                  ``boo ya." \hfill uninformative, high implicature, unformal, &        4k \\
          MNLI &        inference relation &                               ``they renewed inquiries"/``they asked again" \hfill entailment &      391k \\
          STAC &        discourse relation &                                     ``what ?"/``i literally lost" \hfill question-answer-pair &       11k \\
   SwitchBoard &                speech act &                                        ``well , a little different , actually ," \hfill hedge &       19k \\
          MRDA &                speect act &                      ``yeah that 's that 's that 's what i meant ." \hfill acknowledge-answer &       14k \\
 Verifiability &             verifiability &                         ``I've been a physician for 20 years." \hfill verifiable-experiential &        6k \\
    Persuasion &               C/E/P/S/S/R &     ``Co-operation is essential for team work"/``lions hunt in a team" \hfill low specificity &       566 \\
       EmoBank &                     V/A/D &                     ``I wanted to be there.." \hfill low valence, high arousal, low dominance &        5k \\
           GUM &        discourse relation &                               ``do not drink"/``if underage in your country" \hfill condition &        2k \\
          QNLI &        inference relation &                           ``Who took over Samoa?"/``Sykes--Picot Agreement." \hfill entailment &      105k \\
          MNLI &        inference relation &                               ``they renewed inquiries"/``they asked again" \hfill entailment &      391k \\
          STS-B &                similarity &                                           ``a man is running."/``a man is mooing." \hfill dissimilar &        1k \\
          CoLA &  linguistic acceptability &                                                  ``They drank the pub." \hfill not-acceptable &        8k \\
           QQP &                paraphrase &                                   ``Is there a soul?"/``What is a soul?" \hfill Non-duplicate &      364k \\
           RTE &        inference relation &  ``Oil prices fall back as Yukos oil threat lifted"/``Oil prices rise." \hfill not-entailment &        2k \\
          WNLI &        inference relation &              ``The fish ate the worm. It was tasty."/``The fish was tasty." \hfill entailment &       0.6k \\
\bottomrule
\end{tabular}
\caption{Classification datasets considered in our study; $N_{train}$ is the number of training examples}
\label{tab:classification-datasets}
\end{table*}

\subsection{Model}

 \begin{table*}[hbt]
 \footnotesize
     \centering
\begin{tabular}{p{5.6cm}p{5.5cm}ll}
\toprule
                                                                                                                             sentence1 &                                                      sentence2 &          marker &                    sense \\
\midrule
                                                                              {\it every act of god is holy because god is holy .} &     {\it every act of god is loving because god is love .} &       likewise, &     {\textsc similarity} \\
                                                   {\it it gives you a schizophrenic feeling when trying to navigate a web page .} &                        {\it it 's just a bad experience .} &          sadly, &       {\textsc negative} \\
                                                                         {\it the article below was published a few months back .} &                {\it there is all too much truth in this .} &          sadly, &       {\textsc negative} \\
                                                                  {\it i do n't think i can stop with the exclamation marks ! ! !} &                      {\it this could be a problem ! ! ! !} &      seriously, &        {\textsc sarcasm} \\
                                                                          {\it ayite , think of link building as brand building .} &                             {\it there are no shortcuts .} &  unfortunately, &       {\textsc negative} \\
                                                                                           {\it you will seldom meet new people .} &        {\it in medellin you will definitely meet people .} &    in\_contrast, &  {\textsc contradiction} \\
                                                                              {\it if i burn a fingertip , i 'll moan all night .} &                            {\it it did n't look too bad .} &      initially, &  {\textsc contradiction} \\
                                                                              {\it he puncture is about the size of a large pea .} &                         {\it he can see almost no blood .} &      curiously, &  {\textsc contradiction} \\
\bottomrule
\end{tabular}
     \caption{Examples of the Discovery datasets illustrating various relation senses predicted by DiscSense}
     \label{tab:discoverysenseexamples}
 \end{table*}

For our experiments, we make use of BERT  \cite{devlin2018bert}, as a model for relation prediction. 
BERT is a text encoder pre-trained using language modeling having demonstrated state of the art results in various tasks of relation prediction between sentences, which is our use-case. The parameters
are initialized with the pre-trained unsupervised {\it base-uncased}
model and then fine-tuned using the Adam \cite{Kingma2014} optimizer
with 2 iterations on our corpus data, using default
hyperparameters\footnote{\url{https://github.com/huggingface/pytorch-pretrained-BERT/}} otherwise. We ran marker prediction experiments using BERT on both {\it Discovery}
and {\it Wiki20}.

\section{Results}

\subsection{Marker prediction accuracy}
Table \ref{tab:conn-pred} shows the results of the different models on
the prediction of discourse markers. The accuracy of BERT on the {\it
  Discovery} test data is quite high given the large number of classes
(174, perfectly balanced) and sometimes their low semantic
distinguishability. This accuracy is significantly higher than the
score of the Bi-LSTM model in the setup of \newcite{sileo2019discovery}. The BERT
model finetuned on {\it Discovery} outperforms human performance
reported on {\it Wiki20} with no other adaptation than discarding markers not in {\it Wiki20} during inference.\footnote{But note that there is some overlap between training data since BERT pretraining uses Wikipedia text.}
With a further step of fine-tuning (1 epoch on {\it Wiki20}), we
also outperform the best model from \cite{Malmi2018}.  These results
suggest that the BERT+Discovery model captures a significant part of
the use of discourse markers; in the following section, we will apply
it to the prediction of discourse markers for individual categories.

\subsection{Prediction of markers associated to semantic categories}

 For each semantic dataset, consisting of either annotated sentences
 $(s_1,y)$ or annotated sentence pairs ($s_1$,$s_2$,$y$), where $y$ is
 a category, we use the BERT+Discovery model to predict the most
 plausible marker $m$ in each example.  The classification datasets
 thus yield a list of $(y,m)$ pairs.
 Association rules
 \cite{Hipp:2000:AAR:360402.360421} can be used to find interesting
 rules of the form $(m \Rightarrow y)$, or ($y \Rightarrow m$). 
 We discard examples where no marker is predicted, and we discard markers
 that we predicted less than $20$ times for a particular dataset.
Table \ref{tab:associations} shows a sample of markers with the
 highest probability of $P(y|m)$, i.e. the probability of a class
 given a marker. An extended table, which includes a larger sample of significant
 markers for all datasets included in our experiments, is available in appendix A and an even larger, exhaustive table of 2.9k associations is  publicly available.\footnote{\url{https://github.com/synapse-developpement/DiscSense}}
 
 The associations for some markers are intuitively correct ({\it
  likewise} denotes a semantic similarity expected in front of a
paraphrase, {\it sadly} denotes a negative feeling, etc.) and they
display a predictive power much higher than random choices.  Other
associations seem more surprising at first glance, for example, {\it
  seriously} as a marker of sarcasm---although on second thought, it seems a reasonable assumption that {\it seriously} does
not actually signal a serious message, but rather a sarcastic comment
on the preceding sentence. Generally speaking, we notice the same
tendency for each class: our model predicts both fairly obvious
markers ({\it unfortunately} as a marker for negative sentiment, {\it
  in contrast} for contradiction), but equally more inconspicuous
markers (e.g. {\it initially} and {\it curiously} for the same
respective categories) that are perfectly acceptable, even though they
might have been missed by (and indeed are not present in) {\it a
  priori} approaches to discourse marker categorization.  The associations seem to vary across domains (e.g. between CR and SST2) but some markers (e.g. {\it unfortunately}) seem to have more robust associations than others.
  Table \ref{tab:discoverysenseexamples} provides some Discovery samples where the markers are used accordingly.


On a related note, it is encouraging to see that the top markers
predicted on the implicit PDTB dataset are similar to those present in
the more recent English-DimLex lexicon which annotates PDTB categories as senses for discourse markers
\cite{das-etal-2018-constructing}. 
This indicates that our approach
is able to induce genuine discourse markers for discourse categories
that coincide with linguistic intuitions; however, our approach has the
advantage to lay bare less obvious markers, that might easily be
overlooked by an {\it a priori} categorization.


\section{Conclusion}
Based on a model trained for the prediction of discourse markers, we
have established links between the categories of various semantically
annotated datasets and discourse markers. Compared to {\it a priori}
approaches to discourse marker categorization, our method has the
advantage to reveal more inconspicuous but perfectly sensible markers
for particular categories. The resulting associations can
straightforwardly be used to guide corpus analyses, for example to
define an empirically grounded typology of marker use. More qualitative
analyses would be needed to elucidate subtleties in the most
unexpected results. In further work, we plan to use the associations
we found as a heuristic to choose discourse markers whose prediction
is the most helpful for transferable sentence representation learning.

\section{Bibliographical References}
\label{main:ref}

\bibliographystyle{lrec}
\bibliography{lrec2020W-xample}

\begin{thebibliography}{}

\bibitem[\protect\citename{Asher and Lascarides}2003]{asher2003logics}
Asher, N. and Lascarides, A.
\newblock (2003).
\newblock {\em Logics of conversation}.
\newblock Cambridge University Press.

\bibitem[\protect\citename{Asr and Demberg}2012]{Asr2012ImplicitnessOD}
Asr, F.~T. and Demberg, V.
\newblock (2012).
\newblock {Implicitness of Discourse Relations}.
\newblock In {\em COLING}.

\bibitem[\protect\citename{Bowman \bgroup et al.\egroup }2015]{bowman2015large}
Bowman, S.~R., Angeli, G., Potts, C., and Manning, C.~D.
\newblock (2015).
\newblock A large annotated corpus for learning natural language inference.
\newblock {\em arXiv preprint arXiv:1508.05326}.

\bibitem[\protect\citename{Braud and Denis}2016]{braud-denis:2016:EMNLP2016}
Braud, C. and Denis, P.
\newblock (2016).
\newblock {Learning Connective-based Word Representations for Implicit
  Discourse Relation Identification}.
\newblock In {\em Proceedings of the 2016 Conference on Empirical Methods in
  Natural Language Processing}, pages 203--213, Austin, Texas, nov. Association
  for Computational Linguistics.

\bibitem[\protect\citename{Carlson \bgroup et al.\egroup
  }2001]{DBLP:conf/sigdial/CarlsonMO01}
Carlson, L., Marcu, D., and Okurowski, M.~E.
\newblock (2001).
\newblock {Building a Discourse-tagged Corpus in the Framework of Rhetorical
  Structure Theory}.
\newblock In {\em Proceedings of the Second SIGdial Workshop on Discourse and
  Dialogue - Volume 16}, SIGDIAL '01, pages 1--10, Stroudsburg, PA, USA.
  Association for Computational Linguistics.

\bibitem[\protect\citename{Conneau and Kiela}2018]{conneau-kiela-2018-senteval}
Conneau, A. and Kiela, D.
\newblock (2018).
\newblock {S}ent{E}val: An evaluation toolkit for universal sentence
  representations.
\newblock In {\em Proceedings of the Eleventh International Conference on
  Language Resources and Evaluation ({LREC} 2018)}, Miyazaki, Japan, May.
  European Language Resources Association (ELRA).

\bibitem[\protect\citename{Danlos \bgroup et al.\egroup
  }2015]{danlos:hal-01392807}
Danlos, L., Colinet, M., and Steinlin, J.
\newblock (2015).
\newblock {FDTB1: Rep{\'{e}}rage des connecteurs de discours dans un corpus
  fran{\c{c}}ais}.
\newblock {\em Discours - Revue de linguistique, psycholinguistique et
  informatique}, (17), dec.

\bibitem[\protect\citename{Das \bgroup et al.\egroup
  }2018]{das-etal-2018-constructing}
Das, D., Scheffler, T., Bourgonje, P., and Stede, M.
\newblock (2018).
\newblock {Constructing a Lexicon of {\{}English{\}} Discourse Connectives}.
\newblock In {\em Proceedings of the 19th Annual SIGdial Meeting on Discourse
  and Dialogue}, pages 360--365, Melbourne, Australia, jul. Association for
  Computational Linguistics.

\bibitem[\protect\citename{Devlin \bgroup et al.\egroup }2019]{devlin2018bert}
Devlin, J., Chang, M.-W., Lee, K., and Toutanova, K.
\newblock (2019).
\newblock {Bert: Pre-training of deep bidirectional transformers for language
  understanding}.
\newblock In {\em Proceedings of the 2019 Conference of the North American
  Chapter of the Association for Computational Linguistics: Human Language
  Technologies, Volume 1 (Long Papers)}. Association for Computational
  Linguistics.

\bibitem[\protect\citename{Dolan \bgroup et al.\egroup
  }2004]{DBLP:conf/coling/DolanQB04}
Dolan, B., Quirk, C., and Brockett, C.
\newblock (2004).
\newblock {Unsupervised Construction of Large Paraphrase Corpora: Exploiting
  Massively Parallel News Sources}.
\newblock In {\em {\{}COLING{\}} 2004, 20th International Conference on
  Computational Linguistics, Proceedings of the Conference, 23-27 August 2004,
  Geneva, Switzerland}.

\bibitem[\protect\citename{Hipp \bgroup et al.\egroup
  }2000]{Hipp:2000:AAR:360402.360421}
Hipp, J., G{\"{u}}ntzer, U., and Nakhaeizadeh, G.
\newblock (2000).
\newblock {Algorithms for Association Rule Mining {\&}Mdash; a General Survey
  and Comparison}.
\newblock {\em SIGKDD Explor. Newsl.}, 2(1):58--64, jun.

\bibitem[\protect\citename{Jernite \bgroup et al.\egroup }2017]{Jernite2017}
Jernite, Y., Bowman, S.~R., and Sontag, D.
\newblock (2017).
\newblock {Discourse-Based Objectives for Fast Unsupervised Sentence
  Representation Learning}.

\bibitem[\protect\citename{Kingma and Ba}2014]{Kingma2014}
Kingma, D. and Ba, J.
\newblock (2014).
\newblock {Adam: A Method for Stochastic Optimization}.
\newblock {\em International Conference on Learning Representations}, pages
  1--13.

\bibitem[\protect\citename{Knott}1996]{DBLP:phd/ethos/Knott96}
Knott, A.
\newblock (1996).
\newblock {\em {A data-driven methodology for motivating a set of coherence
  relations}}.
\newblock {Ph.D.} thesis, University of Edinburgh, {\{}UK{\}}.

\bibitem[\protect\citename{Liu \bgroup et al.\egroup }2016]{Liu2016ImplicitDR}
Liu, Y.~P., Li, S., Zhang, X., and Sui, Z.
\newblock (2016).
\newblock Implicit discourse relation classification via multi-task neural
  networks.
\newblock {\em ArXiv}, abs/1603.02776.

\bibitem[\protect\citename{Malmi \bgroup et al.\egroup }2018]{Malmi2018}
Malmi, E., Pighin, D., Krause, S., and Kozhevnikov, M.
\newblock (2018).
\newblock {Automatic Prediction of Discourse Connectives}.
\newblock In {\em Proceedings of the 11th Language Resources and Evaluation
  Conference}, Miyazaki, Japan, may. European Language Resource Association.

\bibitem[\protect\citename{Nie \bgroup et al.\egroup }2019]{Nie2017}
Nie, A., Bennett, E., and Goodman, N.
\newblock (2019).
\newblock {D}is{S}ent: Learning sentence representations from explicit
  discourse relations.
\newblock pages 4497--4510, July.

\bibitem[\protect\citename{Pan \bgroup et al.\egroup
  }2018a]{pan-etal-2018-discourse}
Pan, B., Yang, Y., Zhao, Z., Zhuang, Y., Cai, D., and He, X.
\newblock (2018a).
\newblock {Discourse Marker Augmented Network with Reinforcement Learning for
  Natural Language Inference}.
\newblock In {\em Proceedings of the 56th Annual Meeting of the Association for
  Computational Linguistics (Volume 1: Long Papers)}, pages 989--999,
  Melbourne, Australia, jul. Association for Computational Linguistics.

\bibitem[\protect\citename{Pan \bgroup et al.\egroup
  }2018b]{Pan2018DiscourseMA}
Pan, B., Yang, Y., Zhao, Z., Zhuang, Y., Cai, D., and He, X.
\newblock (2018b).
\newblock Discourse marker augmented network with reinforcement learning for
  natural language inference.
\newblock In {\em ACL}.

\bibitem[\protect\citename{Panchenko \bgroup et al.\egroup
  }2017]{Panchenko2017}
Panchenko, A., Ruppert, E., Faralli, S., Ponzetto, S.~P., and Biemann, C.
\newblock (2017).
\newblock {Building a Web-Scale Dependency-Parsed Corpus from Common Crawl}.
\newblock pages 1816--1823.

\bibitem[\protect\citename{Pitler \bgroup et al.\egroup }2008]{pitler2008}
Pitler, E., Raghupathy, M., Mehta, H., Nenkova, A., Lee, A., and Joshi, A.
\newblock (2008).
\newblock {Easily Identifiable Discourse Relations}.
\newblock In {\em Coling 2008: Companion volume: Posters}, pages 87--90. Coling
  2008 Organizing Committee.

\bibitem[\protect\citename{Prasad \bgroup et al.\egroup }2008]{pdtb2.0}
Prasad, R., Dinesh, N., Lee, A., Miltsakaki, E., Robaldo, L., Joshi, A., and
  Webber, B.
\newblock (2008).
\newblock {The Penn Discourse TreeBank 2.0.}
\newblock In Bente Maegaard Joseph Mariani Jan Odijk Stelios Piperidis
  Daniel~Tapias {Nicoletta Calzolari (Conference Chair) Khalid Choukri},
  editor, {\em Proceedings of the Sixth International Conference on Language
  Resources and Evaluation (LREC'08)}, Marrakech, Morocco, may. European
  Language Resources Association (ELRA).

\bibitem[\protect\citename{Roze \bgroup et al.\egroup }2012]{roze12discours}
Roze, C., Danlos, L., and Muller, P.
\newblock (2012).
\newblock {LEXCONN: A French Lexicon of Discourse Connectives}.
\newblock {\em Discours}, (10).

\bibitem[\protect\citename{Sileo \bgroup et al.\egroup
  }2019a]{sileo2019discoursebased}
Sileo, D., de~Cruys, T.~V., Pradel, C., and Muller, P.
\newblock (2019a).
\newblock Discourse-based evaluation of language understanding.

\bibitem[\protect\citename{Sileo \bgroup et al.\egroup
  }2019b]{sileo2019discovery}
Sileo, D., Van De~Cruys, T., Pradel, C., and Muller, P.
\newblock (2019b).
\newblock Mining discourse markers for unsupervised sentence representation
  learning.
\newblock In {\em Proceedings of the 2019 Conference of the North {A}merican
  Chapter of the Association for Computational Linguistics: Human Language
  Technologies, Volume 1 (Long and Short Papers)}, pages 3477--3486,
  Minneapolis, Minnesota, June. Association for Computational Linguistics.

\bibitem[\protect\citename{Stede}2002]{Stede:02}
Stede, M.
\newblock (2002).
\newblock {Di{\{}M{\}}{\{}L{\}}ex: A Lexical Approach to Discourse Markers}.
\newblock In {\em Exploring the Lexicon - Theory and Computation}. Edizioni
  dell'Orso, Alessandria.

\bibitem[\protect\citename{Wang \bgroup et al.\egroup }2019]{wang2018glue}
Wang, A., Singh, A., Michael, J., Hill, F., Levy, O., and Bowman, S.~R.
\newblock (2019).
\newblock {{\{}GLUE{\}}: A Multi-Task Benchmark and Analysis Platform for
  Natural Language Understanding}.
\newblock In {\em International Conference on Learning Representations}.

\bibitem[\protect\citename{Zeyrek and Webber}2008]{zeyrek:discourse:2008}
Zeyrek, D. and Webber, B.
\newblock (2008).
\newblock {A Discourse Resource for Turkish: Annotating Discourse Connectives
  in the {\{}METU{\}} Corpus}.
\newblock In {\em Proceedings of IJCNLP}.

\bibitem[\protect\citename{Zhou \bgroup et al.\egroup }2014]{zhou:chinese:2014}
Zhou, Y., Lu, J., Zhang, J., and Xue, N.
\newblock (2014).
\newblock {Chinese Discourse Treebank 0.5 {\{}LDC2014T21{\}}}.

\end{thebibliography}

\clearpage
\newpage

\appendix
\onecolumn 
\section{DiscSense categories}
\label{appendix}
\begin{center}
    
\normalsize{
\begin{longtable}{llrl}

\toprule
      antecedents &                                 consequents &  support & confidence+prior \\
\midrule
\endhead

   unfortunately, &                                      CR.negative &       66 &     100.0 (21.8) \\
      regardless, &                                      CR.positive &       31 &      96.9 (37.8) \\
         meaning, &                        Cola.not-well-formed &       21 &      48.8 (16.5) \\
      regardless, &                            Cola.well-formed &       23 &      95.8 (39.1) \\
            only, &                            Emergent.against &       24 &       88.9 (8.8) \\
        normally, &                                Emergent.for &       22 &      78.6 (26.8) \\
      separately, &                          Emergent.observing &      148 &      59.2 (20.9) \\
          anyway, &                               EmoBankA.high &       24 &      85.7 (30.0) \\
      originally, &                                EmoBankA.low &       27 &      90.0 (31.8) \\
        together, &                               EmoBankD.high &       20 &      62.5 (23.4) \\
      inevitably, &                                EmoBankD.low &       21 &      91.3 (37.1) \\
            plus, &                               EmoBankV.high &       45 &      90.0 (26.2) \\
           sadly, &                                EmoBankV.low &       36 &      92.3 (34.5) \\
     by contrast, &                              Formality.high &       35 &     100.0 (28.9) \\
            well, &                               Formality.low &       49 &     100.0 (31.0) \\
            this, &                            GUM.circumstance &       24 &       35.3 (7.5) \\
              or, &                               GUM.condition &       31 &       50.0 (5.5) \\
         instead, &                            Implicature.high &       28 &      77.8 (28.3) \\
   by comparison, &                             Implicature.low &       24 &      88.9 (32.2) \\
      nationally, &                        Informativeness.high &       29 &     100.0 (28.3) \\
       seriously, &                         Informativeness.low &       37 &     100.0 (32.8) \\
     in contrast, &                          MNLI.contradiction &     1182 &      74.1 (16.9) \\
         in turn, &                             MNLI.entailment &     7475 &      65.4 (17.0) \\
     for instance &                                MNLI.neutral &      177 &      70.8 (16.9) \\
              so, &                                 MRDA.Accept &       57 &       12.9 (1.5) \\
            well, &                     MRDA.Acknowledge-answer &       85 &       10.3 (1.7) \\
            well, &                       MRDA.Action-directive &       20 &        2.4 (0.7) \\
        actually, &            MRDA.Affirmative Non-yes Answers &       37 &       12.2 (1.5) \\
      personally, &                MRDA.Assessment/Appreciation &       25 &       15.9 (1.9) \\
      especially, &               MRDA.Collaborative Completion &       25 &        7.4 (1.0) \\
          really, &                   MRDA.Declarative-Question &       48 &       11.9 (0.7) \\
          mostly, &                  MRDA.Defending/Explanation &      114 &       62.3 (5.2) \\
        probably, &                   MRDA.Dispreferred Answers &       25 &        1.5 (0.6) \\
          namely, &              MRDA.Expansions of y/n Answers &       37 &       33.6 (4.1) \\
              so, &                          MRDA.Floor Grabber &       56 &       12.7 (2.1) \\
              and &                           MRDA.Floor Holder &       53 &        8.2 (2.4) \\
              and &           MRDA.Hold Before Answer/Agreement &       26 &        4.0 (0.5) \\
      absolutely, &  MRDA.Interrupted/Abandoned/Uninterpretable &       24 &        1.2 (0.6) \\
        probably, &                MRDA.Negative Non-no Answers &       28 &        1.7 (0.4) \\
          though, &                                  MRDA.Offer &       27 &       18.9 (3.4) \\
        honestly, &                          MRDA.Other Answers &       31 &       36.0 (0.6) \\
        actually, &                                 MRDA.Reject &       34 &       11.2 (0.4) \\
        probably, &                            MRDA.Reject-part &       20 &        1.2 (0.2) \\
            also, &                            MRDA.Rising Tone &       66 &       36.7 (3.4) \\
      originally, &                              MRDA.Statement &       20 &      37.0 (10.6) \\
          surely, &                    MRDA.Understanding Check &       26 &       40.6 (2.5) \\
   realistically, &                            MRDA.Wh-Question &       24 &       27.6 (1.1) \\
              or, &                        MRDA.Yes-No-question &       61 &       16.1 (0.8) \\
       elsewhere, &                         MRPC.not-paraphrase &       30 &      81.1 (17.1) \\
       similarly, &                             MRPC.paraphrase &       85 &      87.6 (35.5) \\
              but &                             PDTB.Comparison &       97 &       52.4 (3.8) \\
   by doing this, &                            PDTB.Contingency &       22 &       57.9 (6.7) \\
       currently, &                                 PDTB.Entrel &      212 &       63.5 (7.8) \\
     for instance &                              PDTB.Expansion &      179 &      77.5 (13.5) \\
            then, &                               PDTB.Temporal &       62 &       36.7 (1.4) \\
          rather, &                           PDTB.Alternative &       36 &       25.4 (0.6) \\
            then, &                          PDTB.Asynchronous &       60 &       38.7 (2.4) \\
   by doing this, &                                 PDTB.Cause &       22 &      61.1 (14.8) \\
     additionally &                           PDTB.Conjunction &       47 &      63.5 (12.5) \\
              but &                              PDTB.Contrast &       89 &       61.4 (7.0) \\
     for instance &                         PDTB.Instantiation &      138 &       65.1 (4.8) \\
       elsewhere, &                                  PDTB.List &       41 &       16.2 (1.3) \\
    specifically, &                           PDTB.Restatement &      100 &      67.6 (10.6) \\
      separately, &                             PDTB.Synchrony &       21 &        2.8 (0.7) \\
         moreover &                PersuasivenessEloquence.high &       21 &      46.7 (17.5) \\
           hence, &                 PersuasivenessEloquence.low &       21 &      84.0 (48.6) \\
     undoubtedly, &  PersuasivenessPremiseType.common knowledge &       24 &      85.7 (49.1) \\
     for instance &                PersuasivenessRelevance.high &       25 &      67.6 (41.4) \\
     undoubtedly, &                 PersuasivenessRelevance.low &       21 &      56.8 (27.7) \\
     for instance &              PersuasivenessSpecificity.high &       24 &      82.8 (33.1) \\
     undoubtedly, &               PersuasivenessSpecificity.low &       20 &      87.0 (38.9) \\
     undoubtedly, &                  PersuasivenessStrength.low &       20 &      87.0 (42.9) \\
        likewise, &                             QNLI.entailment &       38 &      74.5 (25.4) \\
      regardless, &                         QNLI.not entailment &       29 &      87.9 (25.5) \\
    collectively, &                               QQP.duplicate &       45 &      68.2 (18.6) \\
           oddly, &                           QQP.not-duplicate &       25 &     100.0 (31.8) \\
     technically, &                              RTE.entailment &       55 &      72.7 (28.1) \\
   by comparison, &                          RTE.not entailment &       29 &      67.4 (27.6) \\
     technically, &                         SICKE.contradiction &       29 &       87.9 (7.8) \\
         in turn, &                            SICKE.entailment &       32 &      64.0 (15.3) \\
       meanwhile, &                               SICKE.neutral &      155 &      92.8 (29.9) \\
   unfortunately, &                                   SST-2.negative &      240 &      96.0 (22.5) \\
      nonetheless &                                   SST-2.positive &      383 &      93.4 (28.4) \\
              so, &                        STAC.Acknowledgement &       40 &       21.3 (5.3) \\
              so, &                 STAC.Clarification question &       23 &       12.2 (1.4) \\
          however &                                STAC.Comment &       91 &       48.7 (5.4) \\
       otherwise, &                            STAC.Conditional &       21 &       25.0 (0.6) \\
          anyway, &                           STAC.Continuation &       52 &       10.4 (3.1) \\
        probably, &                               STAC.Contrast &       76 &       18.9 (1.9) \\
     alternately, &                            STAC.Elaboration &       22 &       59.5 (3.9) \\
      especially, &                            STAC.Explanation &       21 &       12.4 (2.0) \\
          really, &                                 STAC.Q Elab &      147 &       32.5 (2.3) \\
    surprisingly, &                   STAC.Question answer pair &       71 &       89.9 (9.8) \\
         finally, &                                 STAC.Result &      130 &       46.9 (3.1) \\
         finally, &                               STAC.Sequence &       29 &       10.5 (0.4) \\
       currently, &                            STAC.no relation &       50 &      65.8 (10.6) \\
       elsewhere, &                               STS.dissimilar &      516 &      70.0 (14.2) \\
         in turn, &                                  STS.similar &      142 &      60.2 (18.4) \\
       presently, &                              SUBJ.objective &       24 &     100.0 (28.1) \\
   in other words &                             SUBJ.subjective &       61 &     100.0 (28.3) \\
     technically, &                             Sarcasm.notsarcasm &       34 &      72.3 (26.8) \\
       seriously, &                                Sarcasm.sarcasm &      225 &      71.2 (26.7) \\
            well, &       SwitchBoard.Acknowledge (Backchannel) &       30 &        2.8 (0.6) \\
       seriously, &                SwitchBoard.Action-directive &       25 &        4.6 (1.1) \\
            only, &     SwitchBoard.Affirmative Non-yes Answers &       20 &        3.0 (0.8) \\
        actually, &                    SwitchBoard.Agree/Accept &       64 &       17.3 (1.9) \\
        actually, &                    SwitchBoard.Appreciation &       58 &       15.7 (2.4) \\
      especially, &        SwitchBoard.Collaborative Completion &       38 &       10.1 (1.3) \\
          anyway, &            SwitchBoard.Conventional-closing &       82 &       39.4 (1.5) \\
          surely, &     SwitchBoard.Declarative Yes-No-Question &       22 &       20.2 (2.0) \\
              or, &            SwitchBoard.Dispreferred Answers &       24 &        1.7 (0.3) \\
        honestly, &                           SwitchBoard.Hedge &       24 &       19.7 (0.8) \\
              so, &    SwitchBoard.Hold Before Answer/Agreement &       24 &        2.5 (0.6) \\
            only, &         SwitchBoard.Negative Non-no Answers &       43 &        6.4 (0.4) \\
              so, &                   SwitchBoard.Open-Question &       85 &        8.8 (0.8) \\
            well, &                           SwitchBoard.Other &       36 &        3.4 (0.4) \\
              or, &                   SwitchBoard.Other Answers &       25 &        1.8 (0.3) \\
      absolutely, &                       SwitchBoard.Quotation &       88 &        6.2 (1.6) \\
      especially, &                   SwitchBoard.Repeat-phrase &       24 &        6.4 (0.6) \\
              or, &             SwitchBoard.Rhetorical-Question &       48 &        3.4 (0.9) \\
              so, &                       SwitchBoard.Self-talk &       22 &        2.3 (0.2) \\
          really, &        SwitchBoard.Signal-non-understanding &       37 &        5.6 (0.2) \\
         luckily, &           SwitchBoard.Statement-non-opinion &       20 &       71.4 (7.9) \\
      personally, &               SwitchBoard.Statement-opinion &       43 &       20.4 (2.6) \\
         meaning, &           SwitchBoard.Summarize/Reformulate &       26 &        6.9 (1.5) \\
            this, &                 SwitchBoard.Uninterpretable &      158 &       56.0 (9.7) \\
   realistically, &                     SwitchBoard.Wh-Question &       48 &       33.8 (2.9) \\
    incidentally, &                 SwitchBoard.Yes-No-Question &       32 &       78.0 (7.3) \\
  coincidentally, &                  Verifiability.experiential &       20 &       80.0 (8.3) \\
      especially, &              Verifiability.non-experiential &       36 &       39.1 (9.1) \\
           third, &                  Verifiability.unverifiable &       23 &     100.0 (41.3) \\
\bottomrule
\caption[DiscSense mapping sample]{ 
Categories and most associated marker. CR.negative denotes the negative class in the CR dataset. Datasets are described in table \ref{tab:classification-datasets}.  (Supp)ort is the number of examples where the marker was predicted given a dataset. (Conf)idence is the estimated probability of the class given the prediction of the marker i.e. $P(y|m)$. The prior is $P(y)$.
Full version is available at \url{https://github.com/synapse-developpement/DiscSense}
}

\end{longtable}
}
\end{center}


\end{document}